\documentclass[conference]{IEEEtran}

\usepackage{cite}
\usepackage{amsmath,amssymb,amsfonts}
\usepackage{graphicx}
\usepackage{textcomp}
\usepackage{xcolor}

\usepackage{listings}
\usepackage{xspace}
\usepackage[export]{adjustbox}
\usepackage{xcolor}
\usepackage{algorithm}
\usepackage{algpseudocode}
\usepackage{todonotes}
\usepackage{mathtools} 
\usepackage{tikz}
\usepackage{hyperref} 
\usepackage{url}
\usepackage{multirow}
\usepackage{ctable}

\def\BibTeX{{\rm B\kern-.05em{\sc i\kern-.025em b}\kern-.08em
    T\kern-.1667em\lower.7ex\hbox{E}\kern-.125emX}}

\definecolor{dkgreen}{RGB}{0,64,0}
\definecolor{ltgray}{RGB}{245,245,245}
\definecolor{mauve}{RGB}{139,0,139}

\newcommand{\fix}[1]{#1}
\newcommand{\tweakedsim}{\raise.17ex\hbox{$\scriptstyle\mathtt{\sim}$}}

\newcommand\axonn{AxoNN\xspace}
\newcommand\axonnplus{AxoNN}

\newcommand\samo{SAMO\xspace}

\begin{document}

\title{Exploiting Sparsity in Pruned Neural Networks to Optimize Large Model Training}

\author{\IEEEauthorblockN{Siddharth Singh, Abhinav Bhatele}
\IEEEauthorblockA{\textit{Department of Computer Science}\\
\textit{University of Maryland}\\
College Park, Maryland 20742 USA\\
E-mail: ssingh37@umd.edu, bhatele@cs.umd.edu}
}

\maketitle

\begin{abstract}
Parallel training of neural networks at scale is challenging due to significant
overheads arising from communication.  Recently, deep learning researchers have
developed a variety of pruning algorithms that are capable of pruning (i.e.
setting to zero) 80-90\% of the parameters in a neural network to yield sparse
subnetworks that equal the accuracy of the unpruned parent network. In this
work, we propose a novel approach that exploits these sparse subnetworks to
optimize the memory utilization and communication in two popular algorithms for
parallel deep learning namely -- data and inter-layer parallelism. We integrate
our approach into \axonn, a highly scalable framework for parallel deep
learning that relies on data and inter-layer parallelism, and demonstrate the
reduction in communication time and memory utilization.  On 512 NVIDIA V100
GPUs, our optimizations reduce the memory consumption of a 2.7 billion
parameter model by 74\%, and the total communication time by 40\%, thus
providing an overall speedup of 34\% over \axonn, 32\% over DeepSpeed-3D and
46\% over Sputnik, a sparse matrix computation baseline.

\end{abstract}

\begin{IEEEkeywords}
lottery ticket hypothesis, sparse computations, GPUs, parallel deep learning, memory optimizations
\end{IEEEkeywords}

\section{Introduction}
\label{sec:intro}
Deep learning researchers have observed that increasing the size of a neural
network almost always leads to better generalization i.e., accuracies on test
data~\cite{belkin:double-descent}. This has led to the development of neural
architectures with \fix{billions} of parameters~\cite{gpt-3}, which are
naturally trained in parallel on large GPU clusters due to their extreme
compute and memory requirements. The progressive increase in neural network
sizes has necessitated a corresponding increase in the number of GPUs to train
them. However, with increasing GPU counts, communication becomes a significant
bottleneck in the training procedure. Thus, designing algorithms that can
improve the efficiency of training at scale is extremely critical. This will
ensure that we can harness the proven benefits of growing network sizes while
being able to train them in a reasonable amount of time.

The number of parameters in contemporary deep learning models is often in the
tens to hundreds of billions. In their work on the \emph{lottery ticket
hypothesis} (LTH), Frankle et al.~observe empirically that a large fraction of
the parameter set (80-90\%) can be pruned (set to zero) at initialization
without affecting the generalization performance on test
data~\cite{frankle2018the}.  Subsequently, this phenomenon has witnessed great
interest from the deep learning community and several follow up studies have
tried to further refine the hypothesis, propose efficient pruning algorithms
and/or prove it for a broader class of neural network
architectures~\cite{You2020Drawing, prasanna-etal-2020-bert,
structured-pruning-bad, brix-etal-2020-successfully, FrankleD0C20,
Maene2021TowardsUI}. 

\begin{figure}[h]
    \centering
      \includegraphics[width=\columnwidth]{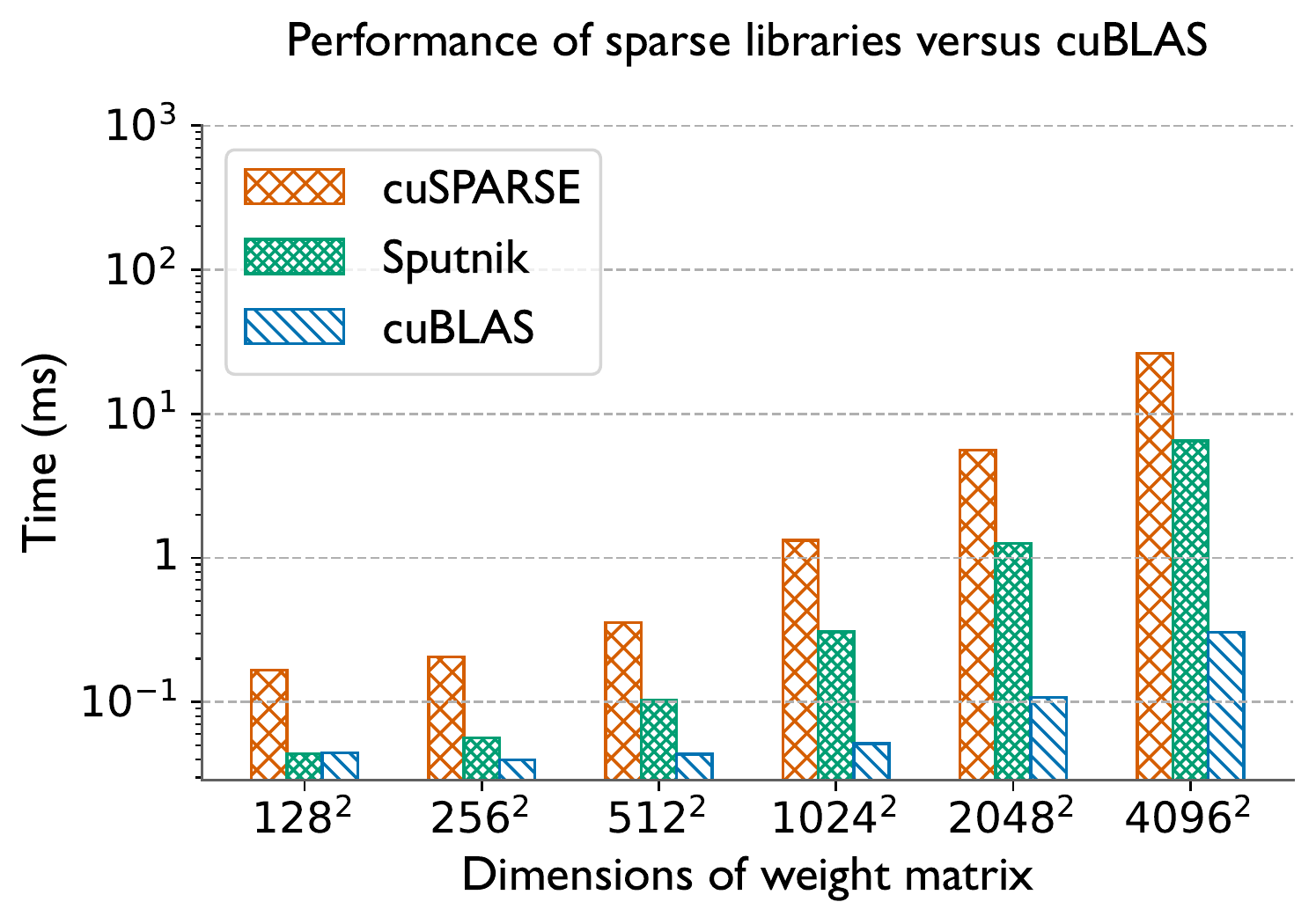}
    \caption{Comparison of the execution times of a fully-connected (FC) layer with a randomly generated, 90\% sparse, square weight matrix in mixed precision.
    FC layers compute a linear transform of their input and are a vital component of various neural network architectures such as transformers\cite{gpt-3}.
    For dense GPU kernels, we use \fix{NVIDIA's cuBLAS}, whereas for sparse GPU kernels, we use \fix{NVIDIA's cuSPARSE} and Sputnik \cite{sputnik}.
    We fix the input batch size to 576 and vary the size of the weight matrix from 
    $128^2$ to $4096^2$.}
    \label{fig:sparse-and-dense}
\end{figure}

Pruning algorithms output extremely sparse subnetworks, which in theory require
significantly fewer number of floating point operations as compared to the
unpruned networks. Several sparse matrix multiplication kernels for GPUs have
been proposed that are specifically optimized for the patterns of sparsities in
these subnetworks~\cite{adaptive-sparse-tiling, sputnik,
sparse-design-principles}. However, in spite of the advancements, these
approaches are significantly slower than \fix{cuBLAS}, a popular library for
dense matrix multiplications on NVIDIA GPUs (used by deep learning frameworks
such as PyTorch and Tensorflow). In Figure~\ref{fig:sparse-and-dense}, we
demonstrate that computing a fully connected layer with 90\% sparsity using
\fix{cuBLAS} (we fill out zeros explicitly in the dense matrix) is
6--22$\times$ faster than using Sputnik~\cite{sputnik}, a state-of-the-art
sparse matrix multiplication library for deep learning workloads. This suggests
that utilizing sparse matrix libraries to improve training performance is
currently infeasible.

Instead of trying to optimize computation times, in this work, we focus on
exploiting sparsity to optimize memory \fix{utilization}, and then exploit the
saved memory to optimize communication.  We demonstrate how our optimizations
can greatly reduce the communication times of two widely used parallel
algorithms for deep learning -- namely inter-layer parallelism (point-to-point
communication) and data parallelism (collective communication).  First, we
propose a novel approach, which we call Sparsity-aware Memory Optimization
(\samo), that provides memory savings of around 66-78\%, while still being
compute efficient.  Through analytical communication models as well as
experiments, we demonstrate that these memory savings can be utilized to reduce
both the message transmission time as well as the pipeline latency (often
called ``bubble" time) of inter-layer parallelism.  Finally, for data
parallelism we only communicate the gradients corresponding to the non-zero
(\fix{or unpruned}) parameters, decreasing the message sizes, and thus
alleviating the bottleneck of expensive all-reduce communication.

To demonstrate the efficacy of our optimizations, we integrate our method in
\axonn~\cite{singh:ipdps2022}, a highly scalable framework for parallel deep
learning that implements an efficient hybrid of data and inter-layer
parallelism. On GPT-3 2.7B~\cite{gpt-3}, a 2.7 billion parameter model, we
demonstrate that \samo reduces the memory \fix{consumption} by 74\% (from 80.16
GB to 20.28 GB)! Then, for a strong scaling run of the same model on 128-512
V100 GPUs of Summit, we successfully exploit the freed up memory to reduce the
portion of batch spent in communication. We show that the absolute reduction in
the communication time accounts for 40\% of \axonn's batch time \fix{on 512
GPUs}. This makes our method a significant 34\% faster than \axonn, 32\% faster
than DeepSpeed-3D~\cite{zero_3D}\fix{,} and 46\% faster that
Sputnik~\cite{sputnik}. Since Sputnik is designed for single GPU executions, we
integrate it in \axonn to run it in parallel.

\noindent We summarize the important contributions of this work below:
\begin{itemize}
\item We present Sparsity-aware Memory Optimization (\samo), a novel method
that exploits recently proposed accuracy-preserving parameter pruning
algorithms in deep learning, to significantly reduce the memory consumption of
neural network training while not sacrificing performance.
\item Through analytical communication models and experiments, we demonstrate
how these memory savings can be utilized to significantly improve the
communication performance of two popular algorithms for parallel deep learning,
data and inter-layer parallelism.
\item We conduct strong scaling experiments using \fix{popular convolution
neural network architectures} and  transformer-based language models with 1.3
to 13 billion parameters on \fix{16 to 2048 GPUs}, and demonstrate significant
improvements in communication times when compared with two highly scalable
parallel deep learning frameworks \axonn and DeepSpeed-3D.
\end{itemize}

\section{Background and Related Work} \label{sec:bg}
Below, we provide background on neural network pruning, sparse matrix
multiplications, and parallel deep learning.

\subsection{Over-parameterization in neural networks}
 
We call a neural network over-parameterized, when its size is extremely large
as compared to the training dataset. It has been empirically observed that the
more over-parameterized a neural network gets, the better it seems to
generalize on a held out test dataset~\cite{belkin:double-descent}. Indeed, the
largest neural networks in deep learning (like the GPT-3~\cite{gpt-3}) are
massively over-parameterized.  This perplexing phenomenon, that cannot be
explained by classical machine learning, has been an active area of research in
deep learning ~\cite{pl-condition, provably, implicit-bias,
belkin:double-descent, rethink, ntk}.

\subsection{Lottery ticket hypothesis}
\label{bg:prune-alg}

Proposed by Frankle et al.~\cite{frankle2018the}., the lottery ticket
hypothesis (LTH) asserts that in a randomly-initialized, over-parameterized
neural network, there exists a subnetwork with one- to two-tenths of the
parameters, which when trained in isolation can match and even improve the
test-set performance of the original neural network. They theorize that in an
over-parameterized network, it is this subnetwork that effectively ends up
being trained, thus preventing over-fitting.  They also present a simple
algorithm to identify this subnetwork. Several follow up studies have tried to
further refine the hypothesis, propose efficient pruning algorithms and/or
prove it for a broader class of neural network
architectures~\cite{You2020Drawing, prasanna-etal-2020-bert,
structured-pruning-bad, brix-etal-2020-successfully, FrankleD0C20,
Maene2021TowardsUI}. In this work, we use You et al.'s algorithm for
pruning~\cite{You2020Drawing}.

\subsection{Accelerated sparse kernels}

NVIDIA's \fix{cuSPARSE} is designed for sparse matrices seen in scientific
applications which have extremely high sparsities ($>$99\%). Therefore, it is
not a suitable candidate for the kinds of sparsities observed in neural network
pruning ($<$90\%). A number of approaches have been proposed that can operate
in these levels of sparsities. Yang et al.~augment merge-based algorithms with
a novel row-based splitting technique to hide memory access
latency~\cite{sparse-design-principles}. Hong et al.~design spMM and sDDMM
(used for backward pass of a fully connected layer) that exploit an adaptive
tiling strategy to reduce global memory access~\cite{adaptive-sparse-tiling}.
Gale et al.~conduct an extensive survey of the sparsity patterns found in
matrices across a variety of deep learning workloads~\cite{state-of-sparsity}.
Using the insights drawn from this study, they design state-of-the-art sparse
kernels for spMM and \fix{sDDMM} for deep learning workloads~\cite{sputnik}. A
number of approaches have been proposed that enforce a certain sparsity
structure. Gray et al.~design GPU kernels for block sparse
matrices~\cite{open-ai-block-sparse}. Chen et al.~propose a novel
column-vector-sparse-encoding for block sparse matrices that provides speedup
over cuBLAS at sparsities as low as 70\% at mixed
precision~\cite{column-vector-sparse-encoding}. Dao et al.~propose a technique
to reduce linear maps to a product of diagonal block sparse matrices and design
kernels for computing their products efficiently~\cite{Dao2020Kaleidoscope,
butterfly-matrices}. 

\subsection{Types of parallelism in deep learning}

Three kinds of parallelism, namely intra-layer, inter-layer and data
parallelism have been proposed in parallel deep learning. Intra-layer
parallelism divides the execution of each layer across GPUs~\cite{megatronlm}.
Inter-layer parallelism assigns a contiguous subset of neural network layers to
each GPU~\cite{narayanan2019pipedream, pipemare, megatronlm-2}. Data
parallelism creates a replica of the entire network on each
GPU~\cite{sc2020zero, zero_infinity}. Usually, frameworks for parallel deep
learning implement a hybrid of data parallelism with one or both of intra- and
inter-layer parallelism~\cite{sc2020zero, zero_3D, megatronlm-2}. For more
details, we refer the reader to Ben-Nun et al.~\cite{bennun2019demystifying}.

\subsection{The \axonn deep learning framework}
\label{sec:axonn-bg}

In this paper, we implement our ideas in a state-of-the-art framework, \axonn,
for parallel deep learning~\cite{singh:ipdps2022}.  \axonn implements a hybrid
of inter-layer and data parallelism. It divides the set of GPUs into
$G_{\mathit{data}}$ groups. Each of these groups operates on an equal sized
shard of the input batch, thus implementing data parallelism. Within each
group, there are $G_{\mathit{inter}}$ GPUs implementing inter-layer
parallelism. To achieve concurrency within this inter-layer parallel groups,
\axonn breaks up the input batch shard into several ``microbatches'' and
processes them in a pipelined fashion. Activations and gradients for a
microbatch are exchanged among neighboring GPUs using point-to-point
communication. As compared to other frameworks, \axonn optimizes this
communication by employing i. asynchronous messaging ii. message driven
scheduling of microbatch operations. The former allows it to overlap
communication with computation, whereas the latter allows it to reduce pipeline
stalls. \axonn supports mixed precision training~\cite{micikevicius2018mixed}
and activation checkpointing~\cite{chen2016training}.

\section{Sparsity-aware Memory Optimization}
In this section, we discuss our approach to exploit sparse networks generated
by pruning methods to significantly reduce the memory consumption of large
model training. We refer to our approach as Sparsity-aware Memory Optimization
(\samo).  We discuss \samo in the context of mixed-precision
training\cite{micikevicius2018mixed}, which is the predominant mode used for
the training of large multi billion-parameter models~\cite{sc2020zero,
bloom176b, megatron-turing-nlg-530b}. However, the optimizations discussed
below are general and can also be applied to single-precision training. 

Mixed-\fix{precision} training involves storing the model parameters and
gradients in both 16-bit (half-precision) and 32-bit (single-precision), and
the optimizer states in 32-bit. The expensive forward and the backward pass are
computed in 16-bit for efficiency, whereas the relatively cheaper optimizer
step is done in 32-bit for accuracy. For more details, we refer the reader to
Micikevicius et al.~\cite{micikevicius2018mixed}. 

Model parameters, gradients and optimizer states are collectively referred to
as the model state~\cite{sc2020zero}. While mixed-\fix{precision} is compute
efficient, storing parameters and gradients in two precisions results in
significantly high memory consumption~\cite{sc2020zero} \fix{(25\% more than
single-precision training)}.  For example, in the case of the widely used
GPT-3\cite{gpt-3}, this adds up to a significant 3.5 TB. \fix{For comparison,
the DRAM capacity of a single V100 GPU on Summit is a mere 16GB.}

Before discussing the details of our approach, we define certain variables as
follows:
\begin{itemize}
    \item $\theta^{16}$ and $\theta^{32}$ -- Network parameters in 16- and 32-bit representation respectively
    \item $\nabla\theta^{16}$ and $\nabla\theta^{32}$ -- Network gradients in 16- and 32-bit representation respectively
    \item $\mathrm{os}$ -- 32-bit optimizer states for the network
    \item $\mathrm{ind}=\bigcup\limits_{i}\mathrm{ind}_{i}$ -- output of a
parameter pruning algorithm, where $\mathrm{ind}_{i}$ stores the indices of the
unpruned (non-zero) parameters for the $i$th layer.
\end{itemize}

Now, we present how \samo can help us in significantly reducing the model state
memory requirements. Note that \samo can be applied only after a neural network
has been sparsified using a pruning algorithm.

\subsection{Performance-preserving model state compression}
\label{sec:state-storage}

We have already seen in Figure \ref{fig:sparse-and-dense} that computing the
forward and backward passes with compressed sparse parameter tensors on GPUs is
not a feasible approach. Thus, a memory optimization that tries to compress
model states will be efficient only if it is able to utilize dense computation
kernels on the GPU. Two important observations about the training process drive
the design of our memory optimizations. First, most of the compute in neural
network training happens in the forward and the backward pass. Second, out of
the various model state tensors discussed previously, the \fix{forward and
backward passes} exclusively use $\theta^{16}$ for computation.  Thus, we do
not compress $\theta^{16}$. This allows us to directly invoke dense computation
kernels on GPUs.  For saving memory, we compress the other model states i.e.,
$\theta^{32}$, $\nabla\theta^{16}$, $\nabla\theta^{32}$, and $os$, which
together still comprise 90\% of the model state memory, even without
\fix{$\theta^{16}$}!  By keeping $\theta^{16}$ in an uncompressed format, we
thus tradeoff a small proportion of the maximum possible memory savings to gain
efficiency in compute.

\subsection{Implementation of compressed storage}

To compress a model state, we convert it to a sparse coordinate (COO) format
using the indices of the unpruned parameters (i.e. $\mathrm{ind}$) output by
the pruning algorithm. However, being 32-bit (32-bit is sufficient for storing
the indices of even the largest models in existence) integers, $\mathrm{ind}$
occupies a non-trivial amount of GPU memory. We tackle this issue in two ways.
First, we note that all of the model state tensors have zeros at the same
indices. Therefore, in our storage scheme, the various COO tensors (i.e.
$\theta^{32}$, $\nabla\theta^{16}$, $\nabla\theta^{32}$, and $os$) share a
common index tensor of non-zero values. Secondly, we convert the index tensors
of any layer \fix{to those of a} hypothetical one-dimensional view. As an
example, say the non-zero indices for a $2\times2$ state tensor are
$[(0,0),(1,1)]$. In a one dimensional view of the same state tensor (i.e.
$4\times1$), the non-zero values are at indices 0 and 3. Thus, we can save
memory by storing only 2 integers (i.e. $[0,3]$), without any loss of
information. In general, for an N-dimensional state tensor, this saves us
N$\times$ memory. Having discussed how the various model states are stored by
\samo to optimize for memory, let us now look at how we compute a batch of data
with this storage schema.

\subsection{Training with \samo}
\label{sec:train-samo}

The computation of a batch in neural network training can be divided into three
phases - the forward pass, the backward pass and the optimizer step. The
forward pass computes the batch loss, the backward pass computes the gradients
of the parameters w.r.t. the batch loss, and the optimizer step updates the
parameters.  Let us \fix{now} look at how these phases are computed efficiently
using \samo. 

\vspace{0.08in}
\noindent{\bf Forward Pass:}
The forward pass of a neural network is done using the half-precision
parameters, $\theta^{16}$. As discussed in Section \ref{sec:state-storage}, we
store $\theta^{16}$ in an uncompressed format with zeros explicitly filled in
for pruned parameters. This allows us to exploit efficient dense computation
kernels for GPUs, like those available in \fix{cuBLAS} and \fix{cuDNN}.  Thus,
the forward pass with \samo is exactly the same as \fix{that in} normal mixed
precision training without \samo. 

\vspace{0.08in}
\noindent{\bf Backward Pass:}
The backward pass also uses $\theta^{16}$ to compute the batch gradients.
Therefore, just like the forward pass, we are able to directly invoke efficient
dense computation kernels. However, in Section \ref{sec:state-storage}, we
discussed that we store the half-precision gradients in a compressed state i.e.
only for the unpruned parameters. Thus, we modify the backward pass to compress
the gradients as soon as they are produced for any layer.  We do this at the
granularity of a layer, and not the entire model, so that we never have to
store the uncompressed gradients for the entire model on the GPU memory.

\vspace{0.08in}
\noindent{\bf Optimizer Step:}
In mixed precision training, the optimizer step consists of three element wise
operations.  The first step involves upscaling $\nabla\theta^{16}$ to
$\nabla\theta^{32}$. The second step is running the optimizer using the
upscaled gradients $\nabla\theta^{32}$  and the optimizer states, $os$ to
update the 32-bit parameters, \fix{$\theta^{32}$}. The final step is to
downscale \fix{$\theta^{32}$} to \fix{$\theta^{16}$}. Let us now see how these
three steps are done with \samo. 

We do the first step of upscaling $\nabla\theta^{16}$ to $\nabla\theta^{32}$
directly on the compressed tensors itself \fix{(as the values for the pruned
parameters are always zero)} using dense computation kernels. Again due to the
same reason, the second step of running the optimizer can be directly computed
on the compressed state tensors using dense kernels. This yields the updated
parameters in 32-bit i.e., \fix{$\theta^{32}$}.  The final step of downcasting
$\theta^{32}$ to $\theta^{16}$ is not straightforward because these tensors are
in a compressed and uncompressed state respectively. To solve this, we first
define a new operation,``expansion'', as the inverse operation of compression.
Essentially, it takes a compressed tensor and the indices of the non-zero
parameters to output the uncompressed version. Now, we do the parameter
down-casting in three steps. First, we delete the now old uncompressed
$\theta^{16}$ from the GPU memory. Then we make a copy  of $\theta^{32}$ in
16-bit. Note that this is essentially the compressed version of our 16-bit
parameters. Finally, we ``expand'' this copy using $\mathrm{ind}$ to obtain the
updated $\theta^{16}$. Thus, the only modification to the optimizer step is an
``expand''  operation in the down-casting step. 

\subsection{Analytical model of memory savings}

In this section, we derive the memory savings as a result of storing model
states with \samo.  We assume that the optimizer of choice is
Adam~\cite{KingmaAdam2014}, which is the go-to optimizer in deep learning for
large model training. Adam stores two optimizer states per parameter.
\fix{However, \samo can be easily extended to work with other optimizers as
well.}

First let us derive the model state memory consumption without pruning. Let
$\phi$ be the total number of parameters in the neural network before pruning.
Now, $\theta^{16}$ and $\nabla\theta^{16}$ take up $2\phi$ bytes each, whereas
$\theta^{32}$ and $\nabla\theta^{32}$ take up $4\phi$ bytes each.  Finally,
$\mathrm{os}$, which are stored in single precision take up $8\phi$ bytes. This
adds up to a total of $20\phi$ bytes (2+2+4+4+8). Let us call this quantity
$M^{default}$.

\begin{figure}[t]
    \centering
      \includegraphics[width=\columnwidth]{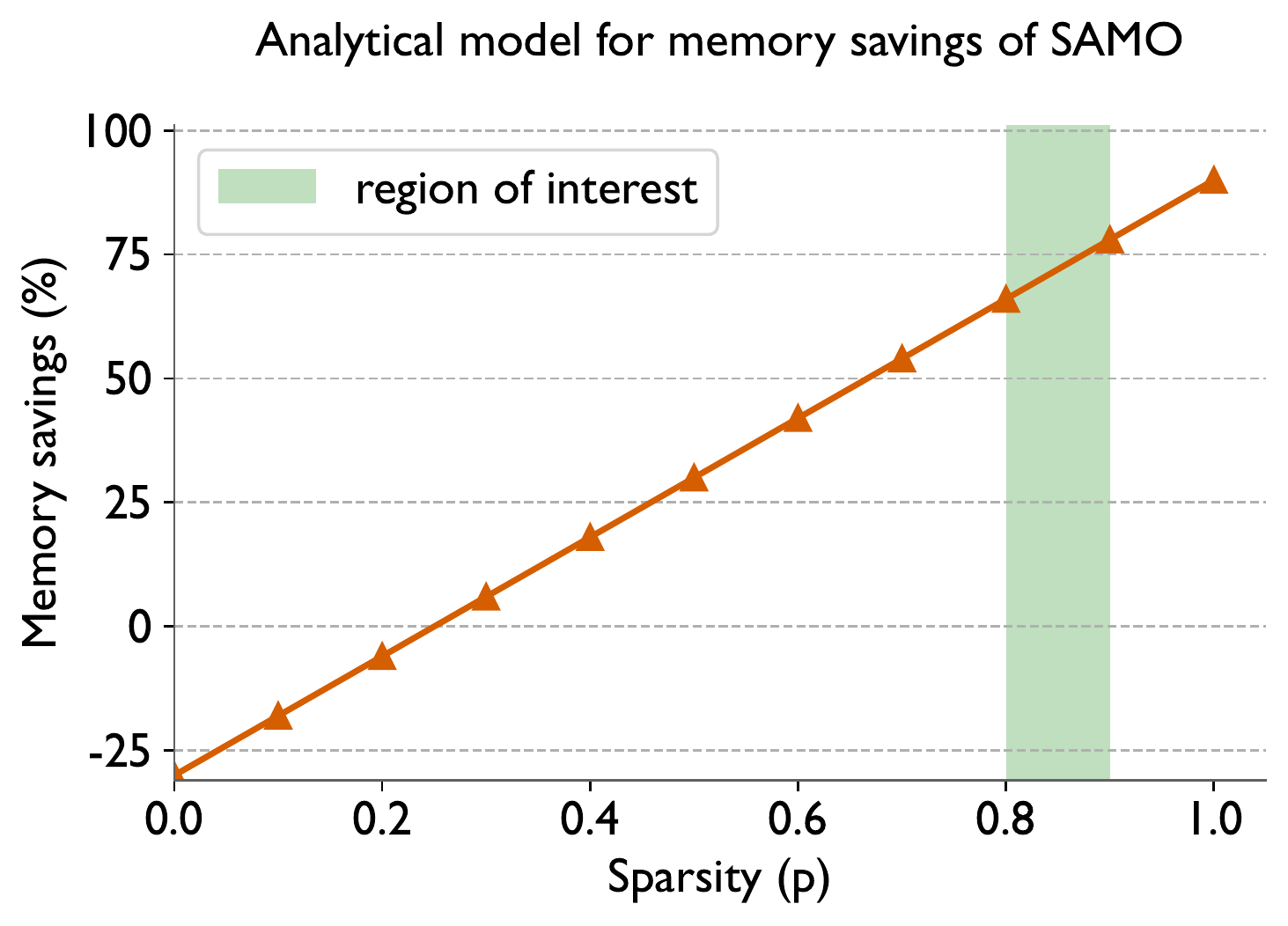}
      \caption{Percentage memory saved by \samo as compared to default mixed-\fix{precision} training. \fix{Sparsity here refers to the 
      proportion of parameters that have been pruned.} \samo can save around 
      66-78\% memory in a range of 0.8-0.9 sparsity, which is typical for most pruning algorithms in deep learning.}
    \label{fig:mem-saved}
\end{figure}

Now, let us assume that we are uniformly pruning $p$ fraction of the parameters
before applying \samo.  This leaves us with $(1-p)\phi$ unpruned parameters.
Let $f=1-p$. We first calculate the memory required to store the compressed
model states i.e. all model states except $\theta^{16}$. For each of these
tensors, we only need to maintain data for $f\phi$ parameters. This adds upto
$18f\phi$ bytes ( $2f\phi$ bytes for $\nabla\theta^{16}$, $4f\phi$ each for
$\theta^{32}$ and $\nabla\theta^{32}$, and another $8f\phi$ for $\mathrm{os}$
). We also maintain a non-zero index per unpruned parameter. In our storage
scheme, each non-zero index is a 32-bit integer. This requires another $4f\phi$
bytes. Storing the uncompressed $\theta^{16}$ state tensor adds a \fix{further}
$2\phi$ bytes. Note that our optimizer step creates a temporary compressed copy
of the half precision parameters at the end of the optimizer step (See Section
\ref{sec:train-samo}).  This adds another $2f\phi$ bytes. Adding everything
together, the total memory consumption of model state storage in bytes is :
\begin{align}
    M^{\samo} & = 18f\phi + 4f\phi + 2\phi + 2f\phi \\
    & = 24f\phi + 2\phi \\
    & = 24(1-p)\phi + 2\phi \\
    & = 20\phi - (24p - 6)\phi \\
    & =  M^{default} - (24p - 6)\phi \label{eqn:memory consumption}
\end{align}
In other words, the absolute amount of \fix{memory savings} that \samo provides
is $(24p - 6)\phi$ bytes, where p is the fraction of parameters that have been
pruned and $\phi$ is the total number of parameters before pruning.  In Figure
\ref{fig:mem-saved}, we plot the percentage memory saved by \samo as compared
to default mixed-\fix{precision} training. We observe that, \samo requires a
minimum sparsity of 0.25 to break even in terms of memory consumption. However,
given that most DL pruning algorithms can comfortably prune 80-90\% of the
parameters, this is not an issue. In this range of sparsities, we observe that
our method saves a significant 66-78\% of memory required to store model
states!

\section{Exploiting \samo for Improving Parallel Training Performance} \label{sec:imp_perf}
When computing on a single GPU, \samo simply reduces memory consumption with
some overheads in the backward pass (compression of gradients) and the
optimizer step (expansion of parameters). Hence, when training on a single GPU,
\samo does not lead to any performance improvements. This is because the total
number of floating point operations in the forward and backward pass is
unchanged (since we still compute in dense).  In this section, we discuss how
parameter pruning and \samo can be used to optimize the performance of
\fix{multi-GPU training}.

The main performance bottleneck in parallel neural network training is
communication. GPUs perform computation on data at a much faster rate than that
of data communication between them on modern HPC interconnects. This problem is
only exacerbated when training larger models, which require a correspondingly
larger number of GPUs on a cluster. Thus, designing algorithms that can
decrease the amount of communication can greatly benefit parallel deep
learning. We now discuss how the application of \samo on a pruned neural
network can reduce communication in parallel training. As discussed in Section
\ref{sec:bg}, we use \axonn, which implements a hybrid of inter-layer
parallelism (point-to-point communication) and data parallelism (collective
communication), to demonstrate the efficacy of our optimizations.

\subsection{Optimizing collective communication in data parallelism}

First, let us see how our optimizations can decrease the overhead of collective
communication in the data parallel phase. After the end of the forward and
backward pass, \axonn synchronizes the local gradients of each GPU via an
all-reduce. In Section \ref{sec:state-storage}, we showed how \samo stores the
16-bit gradients in a compressed format i.e.~only for the unpruned parameters.
This allows us to reduce the size of collective communication messages by
directly invoking \axonn's all-reduce calls on the compressed tensor. This
leads to a significant reduction in the collective communication time. 

\subsection{Optimizing point-to-point communication in inter-layer parallelism}

As described in Section~\ref{sec:bg}, \axonn implements a hybrid of inter-layer
and data parallelism by dividing the work among $G_{\mathrm{inter}} \times
G_{\mathrm{data}}$ GPUs. When \samo is used to reduce the memory required for
training a neural network, we can reduce the number of GPUs required to deploy
a single instance of the neural network i.e.~decrease $G_{\mathrm{inter}}$.
This can allow us to use more GPUs for data parallelism, and increase
$G_{\mathrm{data}}$. A reduced $G_{\mathrm{inter}}$ has the effect of
decreasing the time spent in point-to-point communication thereby increasing
the efficiency of inter-layer parallelism. We now provide a proof for this
claim. We use the following notations:
\begin{itemize}
    \item $B$ - Batch size
    \item $\mathrm{mbs}$ - The size of each microbatch
    \item $G$ - Number of GPUs
    \item $t_{f}$ - Time spent in computation on a microbatch of
size $\mathrm{mbs}$ during the forward pass through the entire model
    \item $t_{b}$ - Time spent in computation on a microbatch of
size $\mathrm{mbs}$ during the backward pass through the entire model
\end{itemize}
Note that $t_{f}$ and $t_{b}$ do not take the point-to-point communication cost
into account. They just denote the compute time for the forward and backward
pass across all the layers.

\begin{figure}[h]
    \centering
      \includegraphics[width=\columnwidth]{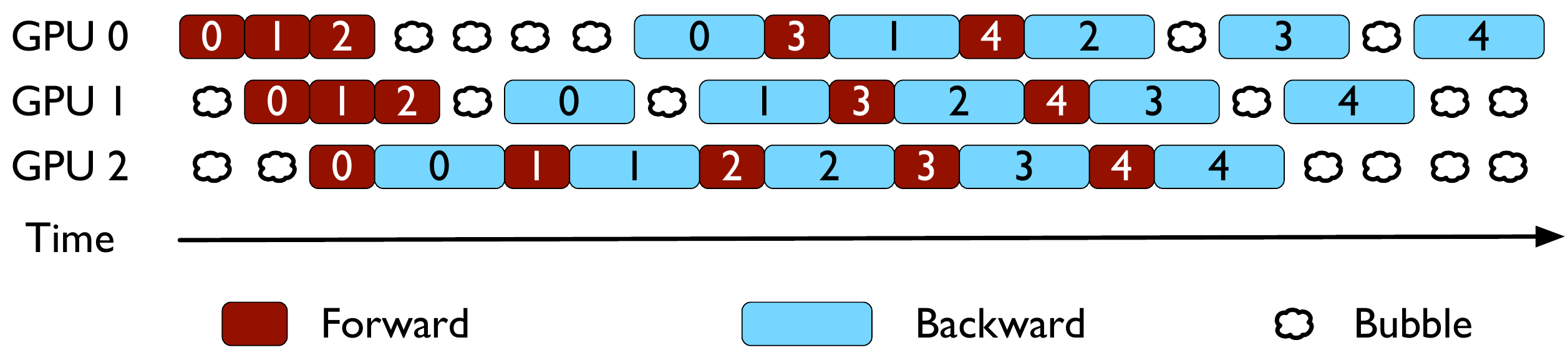}
      \caption{\fix{Illustration} of how a batch is computed in inter-layer parallelism on three GPUs ($G_{\mathrm{inter}}=3$).
      In this example, we have divided the input batch into 5 microbatches (numbered 0 to 4). The red and blue colors
      denote forward and backward passes of microbatches respectively. We assume that the forward pass takes one unit of time 
      and the backward pass takes two units of time. We observe that on each GPU, the pipeline bubble time accounts for 6 units, 
      which equals the time to do $G_{\mathrm{inter}}-1=2$ forward passes and $G_{\mathrm{inter}}-1=2$ backward passes.}
    \label{fig:pipelining}
\end{figure}

The time spent in point-to-point communication can be divided into two parts:
the bubble time and the transmission time. A GPU experiences a pipeline bubble
when there aren't enough microbatches in the pipeline to keep all of the GPUs
busy. As shown in Figure \ref{fig:pipelining}, different GPUs experience
pipeline bubbles at different points in time. But a common theme is that
pipeline bubbles occur towards the beginning and end of the computation of a
batch. We define the transmission time as the total time spent in sending
messages in the pipeline. 

Let $t_{\mathrm{bubble}}$ and $t_{\mathrm{send}}$ denote the bubble time and
transmission time respectively. Narayanan et al.~\cite{megatronlm-2} show that
$t_{\mathrm{bubble}}$ equals the time it takes to complete forward and backward
passes for $G_{\mathrm{inter}}-1$ microbatches on any GPU.  We can also see
this in Figure \ref{fig:pipelining}, wherein we observe that the bubble time
for a pipeline with $G_{\mathrm{inter}}=3$ equals the time to do two forward
and two backward passes. Assuming uniform distribution of compute, the time to
complete the forward and backward pass of a microbatch on a single GPU is
$\frac{t_{f}+t_{b}}{G_{\mathrm{inter}}}$.

Thus, the bubble time can be calculated as,
\begin{align} 
    t_{\mathrm{bubble}} & = (G_{\mathrm{inter}}-1) \times (\frac{t_{f}+t_{b}}{G_{\mathrm{inter}}}) \\
    & =  (t_{f}+t_{b}) \times (1 - \frac{1}{G_{\mathrm{inter}}}) \label{eq:bubble}
\end{align}

Now, taking the derivative of $t_{\mathrm{bubble}}$ with  $G_{\mathrm{inter}}$,
we can show that the pipeline bubble time is a monotonically increasing
function of $G_{\mathrm{inter}}$:
\begin{align}
    \frac{\partial t_{\mathrm{bubble}}}{\partial G_{\mathrm{inter}}} & = \frac{t_{f}+t_{b}}{G^{2}_{\mathrm{inter}}} > 0 \label{eqn:bubble-grad}
\end{align}

Since \samo can help in decreasing $G_{\mathrm{inter}}$ via its memory savings,
we can conclude that it can be used to optimize the pipeline bubble time. Note
that in Equation \ref{eqn:bubble-grad}, we observe that the gradient w.r.t.
$G_{\mathrm{inter}}$ is inversely proportional to its square. Thus, with a
progressive increase in model size (which entails a corresponding increase in
$G_{\mathrm{inter}}$), we expect diminishing returns in the bubble time
improvement.

The transmission time $t_{\mathrm{send}}$ is proportional to the number of
messages sent and received by each GPU. Each GPU sends and receives four
messages per microbatch, two each in the forward and backward passes. Let us
now derive the total number of microbatches each GPU computes on. First, \axonn
divides the input batch into $G_{\mathrm{data}}$ shards, one for each
inter-layer parallel group. Next, each inter-layer parallel group breaks this
batch shard into microbatches of size $\mathrm{mbs}$. These microbatches are
processed by every GPU in the inter-layer parallel group. Thus the total number
of microbatches computed upon by every GPU is
$\frac{B}{G_{\mathrm{data}}\times\mathrm{mbs}}$. Thus, we can express
$t_{\mathrm{send}}$ as,
\begin{align} 
    t_{\mathrm{send}} & \propto  4 \times \frac{B}{\mathrm{mbs} \times G_{\mathrm{data}}}  \\
                       & \propto  4 \times \frac{B}{\mathrm{mbs}} \times \frac{G_{\mathrm{inter}}}{G}  (\because  G_{\mathrm{inter}} \times G_{\mathrm{data}} = G)\label{eq:trans}
\end{align}

Taking the derivative of Equation \ref{eq:trans} w.r.t.~$G_{\mathrm{inter}}$
shows that $t_{\mathrm{send}}$ is a monotonically increasing function of
$G_{\mathrm{inter}}$:
\begin{align} \label{eq:p2p-imp}
    \frac{\partial t_{\mathrm{send}}}{\partial G_{\mathrm{inter}}} & \propto \frac{B}{\mathrm{mbs} \times G} > 0
\end{align}

Hence, we can see that using \samo to decrease $G_{\mathrm{inter}}$ can also
help us decrease the transmission time for point-to-point communication in
inter-layer parallelism.  Thus, we have shown how the memory optimizations in
\samo can be exploited to reduce the collective communication pertaining to
data parallelism and point-to-point communication pertaining to inter-layer
parallelism respectively. Later, in Section \ref{sec:res}, we provide
performance profiles that demonstrate reduction in communication times as
empirical evidence for the claims we have made in this section.

\section{Experimental Setup} \label{sec:setup}
\begin{figure*}[t]
  \centering
    \includegraphics[width=\columnwidth]{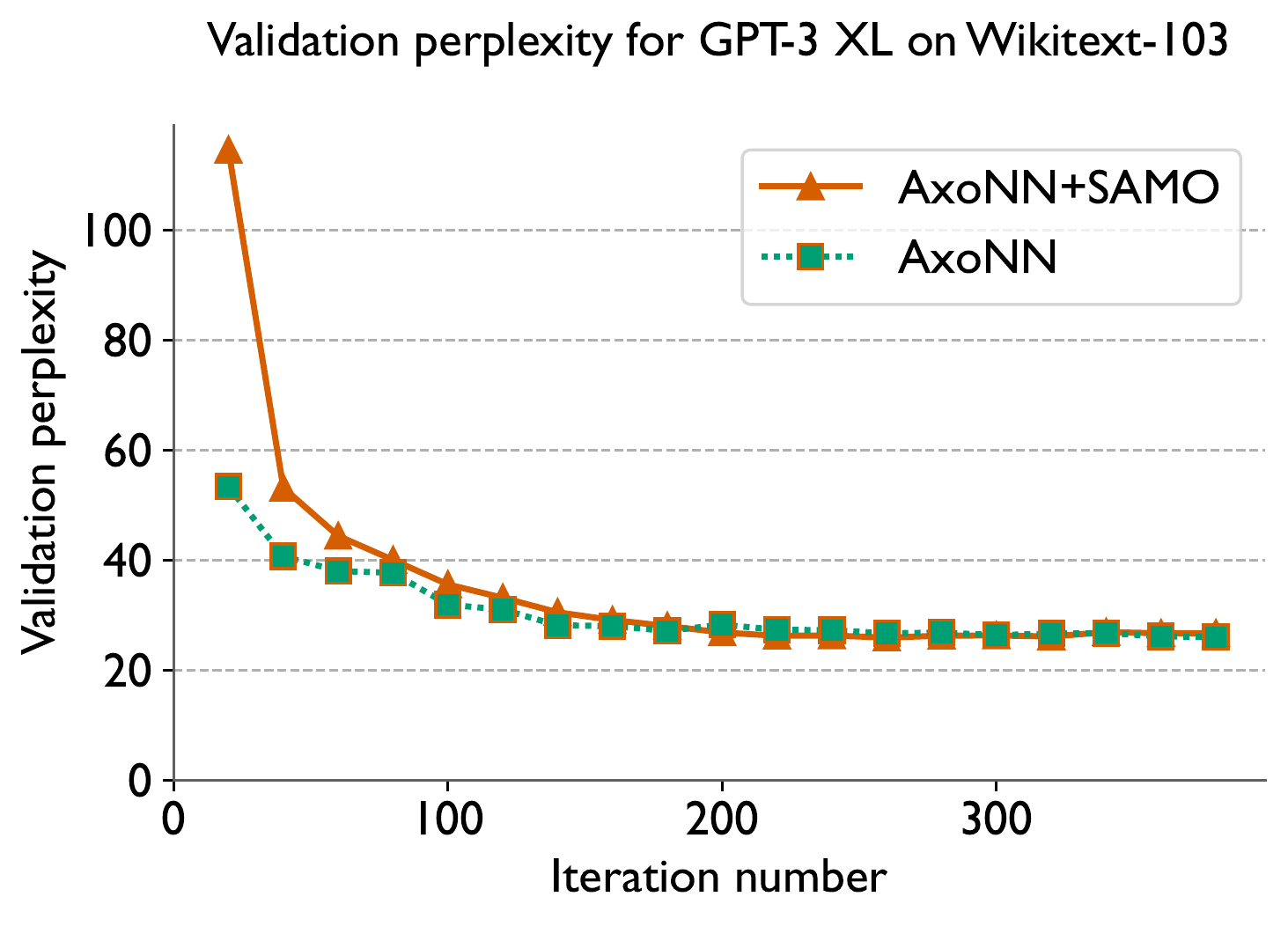}
    \includegraphics[width=\columnwidth]{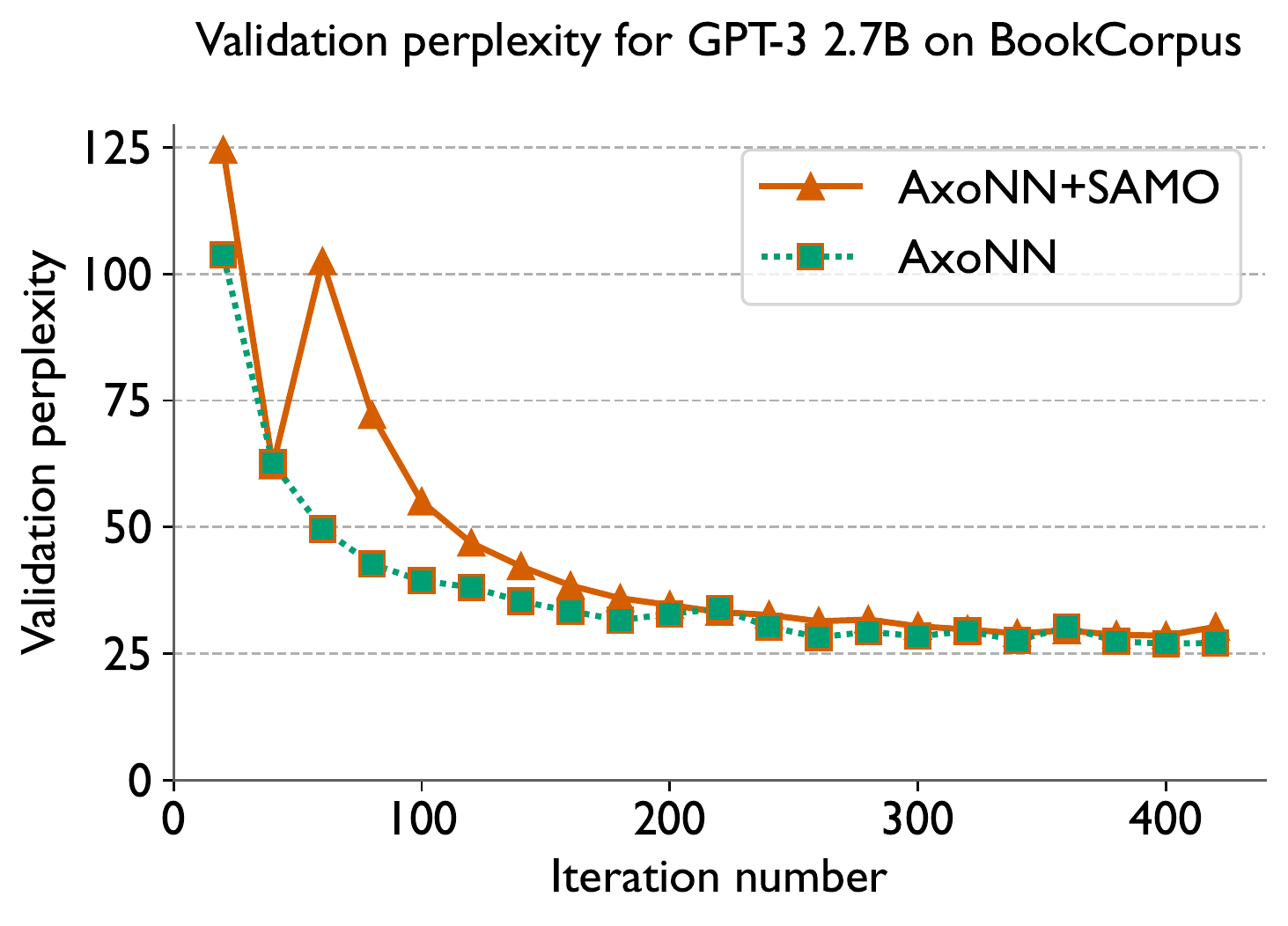}
    \vspace{-0.05in}
    \caption{Validation perplexities for GPT-3 XL (left) and GPT-3 2.7B (right) on 64 and 128
    GPUs of Summit respectively. For \axonnplus+\samo, we prune both models to a
    sparsity of 90\% using~\cite{You2020Drawing}. We use
    the same hyperparameters as Brown et al.~\cite{gpt-3} and train on the
    Wikitext-103~\cite{wikitext-103} and BookCorpus datasets~\cite{book-corpus}.}
    \vspace{-0.03in}
  \label{fig:stat-eff}
\end{figure*}

In this section, we provide details of the empirical experiments that we
conducted to demonstrate the benefits of our optimizations. As discussed in
Section~\ref{sec:bg}, we used \axonn~\cite{singh:ipdps2022} for parallelizing
the training process. We first validate the statistical efficiency of our
implementation by training two neural networks to completion at a sparsity of
0.9. For pruning, we use You et al.'s ``Early-Bird Tickets" pruning
algorithm~\cite{You2020Drawing}. Then, we study the performance of \fix{two
convolution neural networks (VGG-19~\cite{vgg16-iclr} and
WideResnet-101~\cite{wide-resnet}) and} four GPT-style transformer models from
Brown et al.~\cite{gpt-3} under a strong scaling setup to demonstrate the
hardware efficiency of our approach. We use the Oak Ridge National Laboratory's
Summit supercomputer to run our experiments.  Summit has two POWER9 CPUs and
six 16 GB NVIDIA V100 GPUs per node. Each CPU is connected to 3 GPUs via
NVlink. The intra-node bandwidth, inter-node bandwidth, and the peak
half-precision throughput are 50 GB/s, 12.5 GB/s and 125 Tflop/s per GPU
respectively. 

\subsection{Description of neural networks and hyperparameters}

Table~\ref{tab:setup-perf} lists the set of neural networks and the
corresponding hyperparameters used in this study. \fix{VGG-19~\cite{vgg16-iclr}
and WideResnet-101~\cite{wide-resnet} are two convolutional neural network
(CNN) architectures widely used in computer vision. GPT-3~\cite{gpt-3}, a
variant of the transformer architecture~\cite{vaswani2017attention}, is
extremely popular in natural language processing for causal language modeling.}
For each model, we use the same hyperparameters (batch size, sequence length,
learning rate schedules, gradient clipping, l2 regularization and optimizer
hyperparameters) as used by the authors. \fix{We use SGD (with
momentum~\cite{sgd-momentum}) and the AdamW~\cite{adamw} optimizer for training
the CNN and GPT-3 models respectively}. We use MegatronLM's highly optimized
kernels to implement the GPT-3 models~\cite{megatronlm}. \fix{For the
convolution neural networks, we use implementations provided by the torchvision
library\footnote{https://pytorch.org/vision/stable/index.html}.}

\begin{table}[h]
  \centering
  \caption{\label{tab:setup-perf}List of neural networks used in this study.
  For each model, we list the minimum and maximum number of GPUs used in our
  strong scaling runs. We choose the minimum and maximum GPU counts such that
  the ratio of batch size to number of GPUs is 4 and 1 respectively.}
  \begin{tabular}{lrrr}
  \toprule
  Neural Network & \# Parameters & Batch Size & No.~of GPUs      \\ \midrule
  \fix{WideResnet-101~\cite{wide-resnet}} & \fix{126.89M}    & \fix{128}        & \fix{16--128}   \\
  \fix{VGG-19~\cite{vgg16-iclr}}         & \fix{143.67M}    & \fix{128}       & \fix{16--128}   \\
  GPT-3 XL~\cite{gpt-3}      & 1.3B       & 512        & 64--512   \\
  GPT-3 2.7B~\cite{gpt-3}     & 2.7B       & 512        & 64--512    \\
  GPT-3 6.7B~\cite{gpt-3}     & 6.7B       & 1024       & 128--1024 \\ 
  GPT-3 13B~\cite{gpt-3}      & 13B        & 2048       & 256--2048 \\
  \bottomrule
  \end{tabular}
\end{table}

We profile the neural networks listed in Table \ref{tab:setup-perf} under a
strong scaling setup to demonstrate the efficacy of our optimizations. For
every model, we choose the minimum and maximum GPU counts such that the ratio
of batch size to number of GPUs is 4 and 1 respectively. For a given model, we
fix the batch size irrespective of the GPU count. This is because while
increasing the batch size leads to better performance, it also degrades the
quality of convergence~\cite{keskar2017on}. Under a strict definition of strong
scaling, the final answer should be the same irrespective of the number of
GPUs. Therefore, it is important to keep the global batch size fixed. For our
approach, we prune the networks to a sparsity of 90\% using You et al.'s
``Early Bird Ticket'' algorithm~\cite{You2020Drawing}.

To ensure the correctness of our optimizations, we train GPT-3 XL and GPT-3
2.7B to completion on the Wikitext-103 dataset~\cite{wikitext-103} and the
BookCorpus dataset~\cite{book-corpus} respectively. We present the validation
perplexity curves for \axonn and \axonnplus+\samo. Again, we use a sparsity of
90\% and the same pruning algorithm as the strong scaling runs. \fix{The
purpose of this experiment to ensure that our proposed optimizations work
correctly in an end-to-end fashion in combination with a pruning algorithm.
Since this is a sanity check, the datasets we have used are much smaller than
what are typically used to train large GPT-3 style language models. We
highlight prior work by Samar et al.~\cite{cerebas-gpt-pruning}, who have
successfully pruned GPT-3 style language models upto 90\% on a much larger and
challenging dataset (Pile~\cite{the_pile}.)}

\subsection{Choice of frameworks}

We integrate our optimizations in \axonn~\cite{singh:ipdps2022} and refer to it
as ``\axonnplus+\samo''. We use \axonn and
DeepSpeed-3D~\cite{zero_3D,sc2020zero} as baselines for dense computations.
DeepSpeed-3D implements a hybrid of data, inter-layer and intra-layer
parallelism for parallel model training. Their data parallel implementation
uses the ZeRO optimizer to shard optimizer state memory across data parallel
ranks~\cite{sc2020zero}. They use MegatronLM's implementation of intra-layer
parallelism of transformers~\cite{megatronlm}. DeepSpeed-3D has been used to
train some of the largest neural networks till date such as
Bloom-176B~\cite{bloom176b} and Megatron-Turing NLG
530B~\cite{megatron-turing-nlg-530b}. Finally, we integrate Sputnik
~\cite{sputnik} in \axonn to create a sparse matrix multiplication baseline.
\fix{Note that Sputnik does not support sparse convolutions, so we do not
implement the convolution architectures in Table \ref{tab:setup-perf} using
Sputnik.} We build our baselines using CUDA 11.0, PyTorch 1.12.0, NCCL 2.8.4,
GCC 9.1.0 and Spectrum-MPI 10.4.0.3.

\subsection{Evaluation metrics}

For the statistical efficiency experiments, we report perplexity on the
validation split of the training dataset. Perplexity is defined as the
exponential of the cross entropy loss. \fix{For our strong scaling runs, we
report the average iteration time i.e. time to train on a single batch of input
data. We do this by training for 100 batches and averaging the time of the last
90.}

For \fix{the transformer models}, we also calculate the percentage of peak
half-precision flop/s. To do this, we use Narayanan et al.'s formula to derive
the total number of floating point operations in a batch~\cite{megatronlm-2} of
a transformer model and divide it by the average batch time over 100 training
batches to derive the flop/s. Finally, we divide this quantity by 125 Tflop/s
(the peak half-precision flops per GPU on Summit) and the number of GPUs to
obtain the percentage of peak half-precision throughput.

Since Sputnik is a sparse matrix multiplication library, it only computes a
fraction of the flops that the other dense computation frameworks compute. For
instance at a sparsity of 90\%, it would only compute 10\% of the flops. For a
fair comparison, we assume the same number of flops for Sputnik as the dense
computation frameworks while using the time spent computing the sparse kernels.

\section{Results} \label{sec:res}
We now present the results of the empirical experiments outlined in Section
\ref{sec:setup}.

\subsection{Statistical efficiency}

We verify the statistical efficiency of \axonnplus+\samo by training GPT-3
XL~\cite{gpt-3} and GPT-3 2.7B~\cite{gpt-3} to completion at a sparsity of
90\%. We use You et al.'s algorithm~\cite{You2020Drawing} to prune a neural
network for \axonnplus+\samo.  Figure \ref{fig:stat-eff} illustrates the
results of this experiment. We observe that (1) the final validation
perplexities for the pruned networks trained with \axonnplus+\samo match those
of the unpruned network trained with \axonn and (2) both \axonn and
\axonnplus+\samo reach the final validation perplexities in similar number of
iterations. This verifies the correctness of our implementation.

\subsection{Strong scaling performance}

\begin{figure*}[h]
  \centering
    \includegraphics[width=\columnwidth]{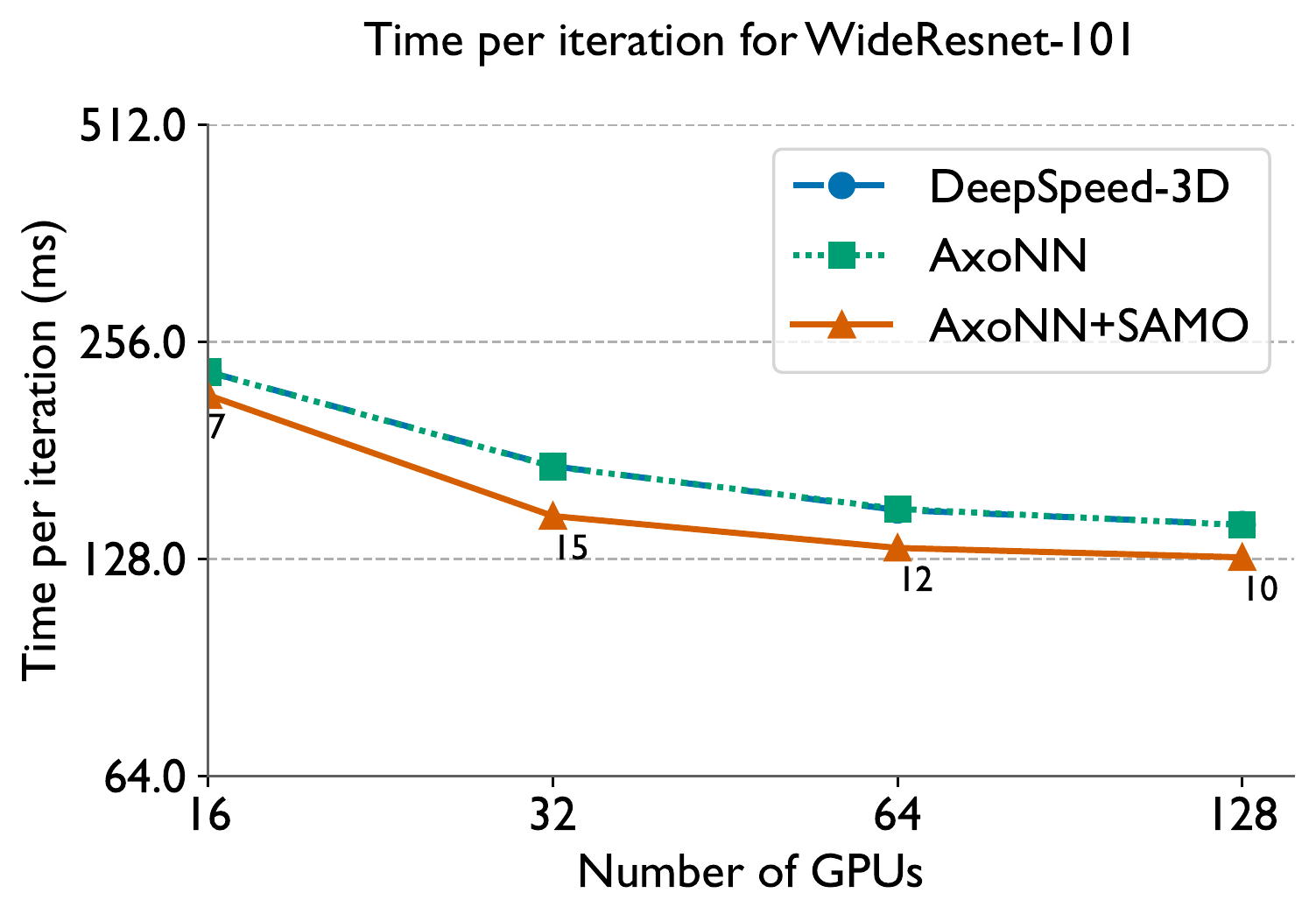}
    \includegraphics[width=\columnwidth]{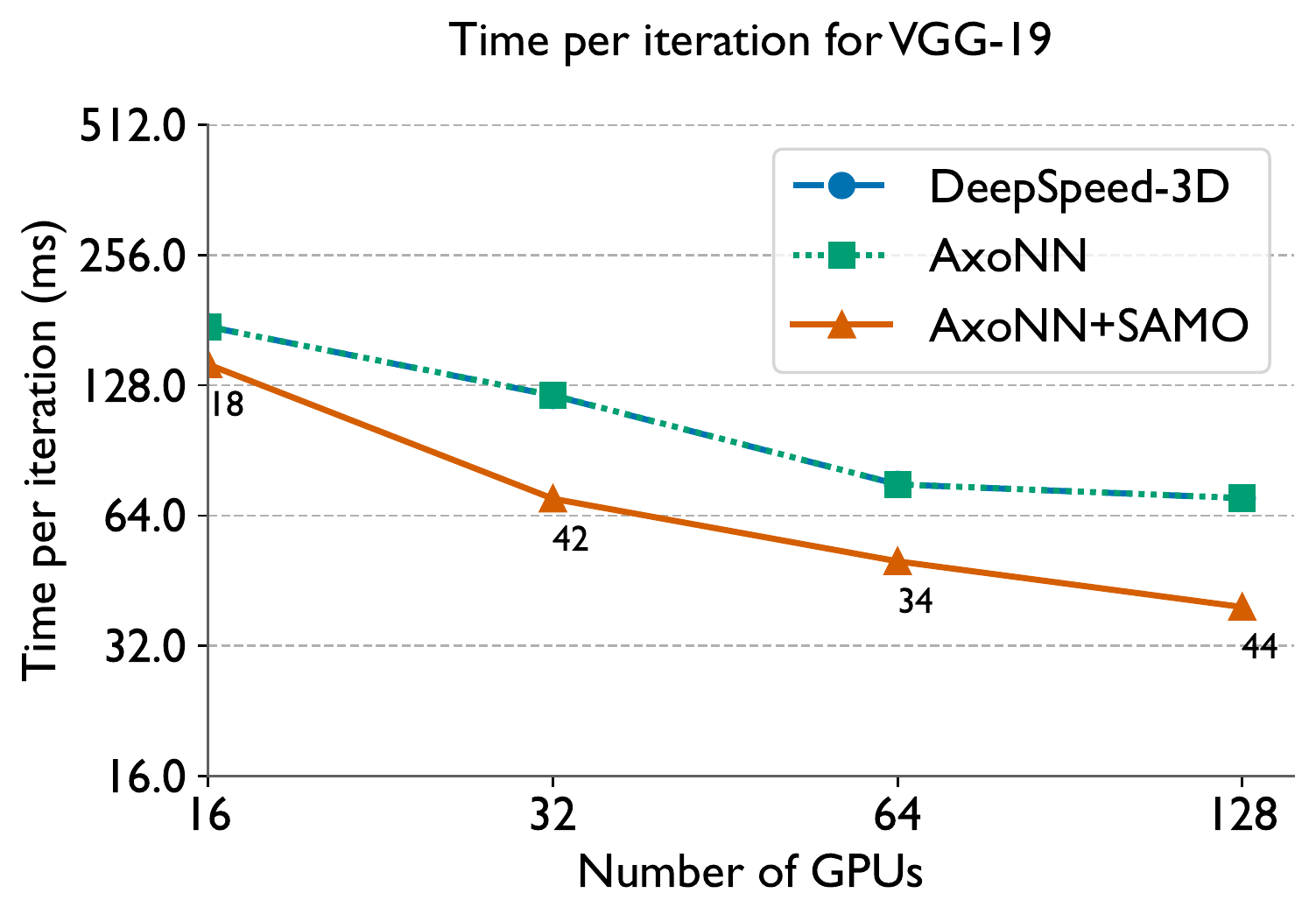}
    \caption{\fix{Time per iteration (batch time) for a strong scaling
  study of WideResnet-101 (left) and VGG-19 (right) on Summit. We prune the models to a sparsity of 90\% 
  for \axonnplus+\samo (see Table \ref{tab:setup-perf} for batch sizes).
  We annotate \axonnplus+\samo's line with its percentage speedup over \axonn.}}
  \label{fig:hardware-eff-cnn}
\end{figure*}

\begin{figure*}[h]
  \centering
    \includegraphics[width=\columnwidth]{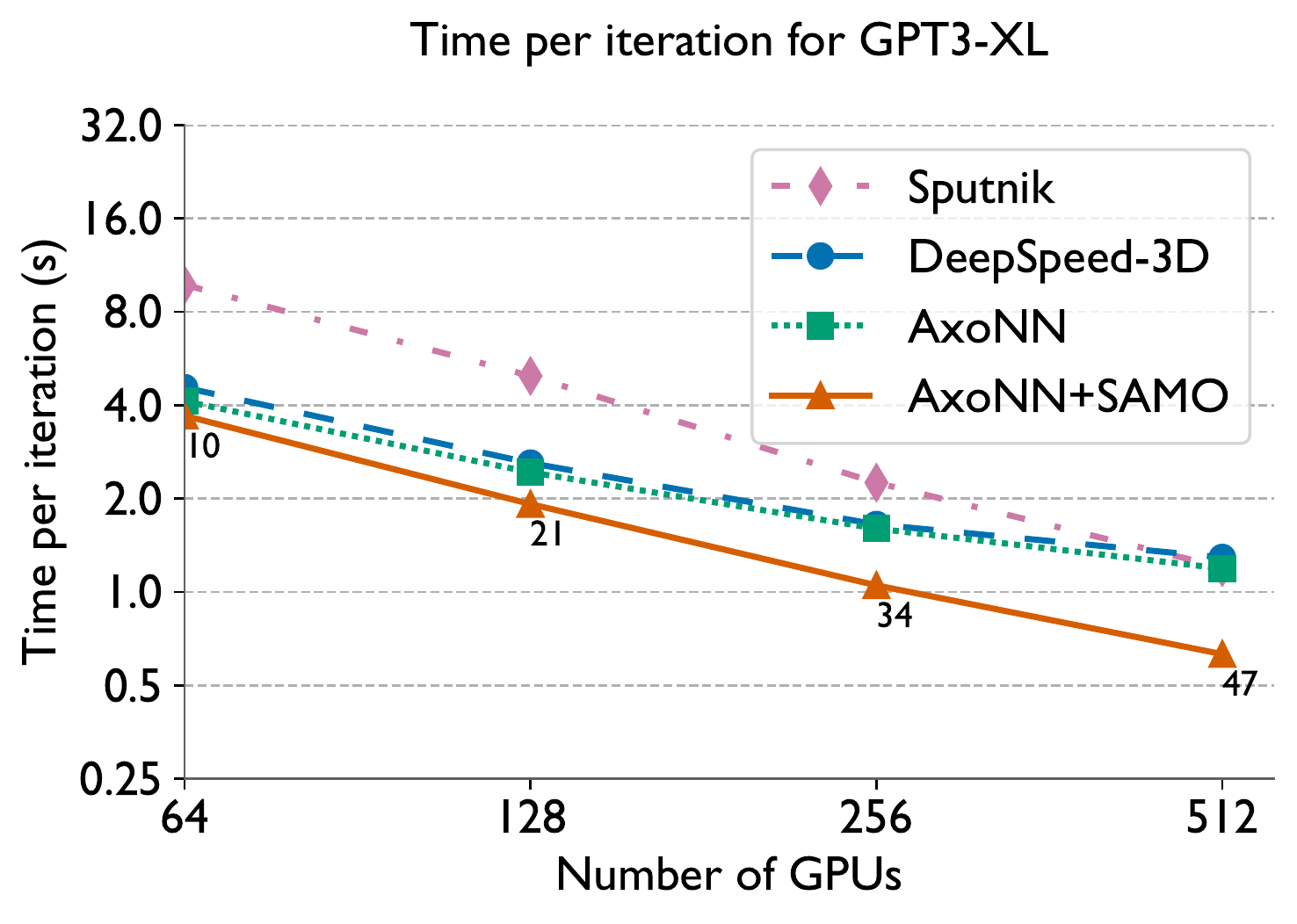}
    \includegraphics[width=\columnwidth]{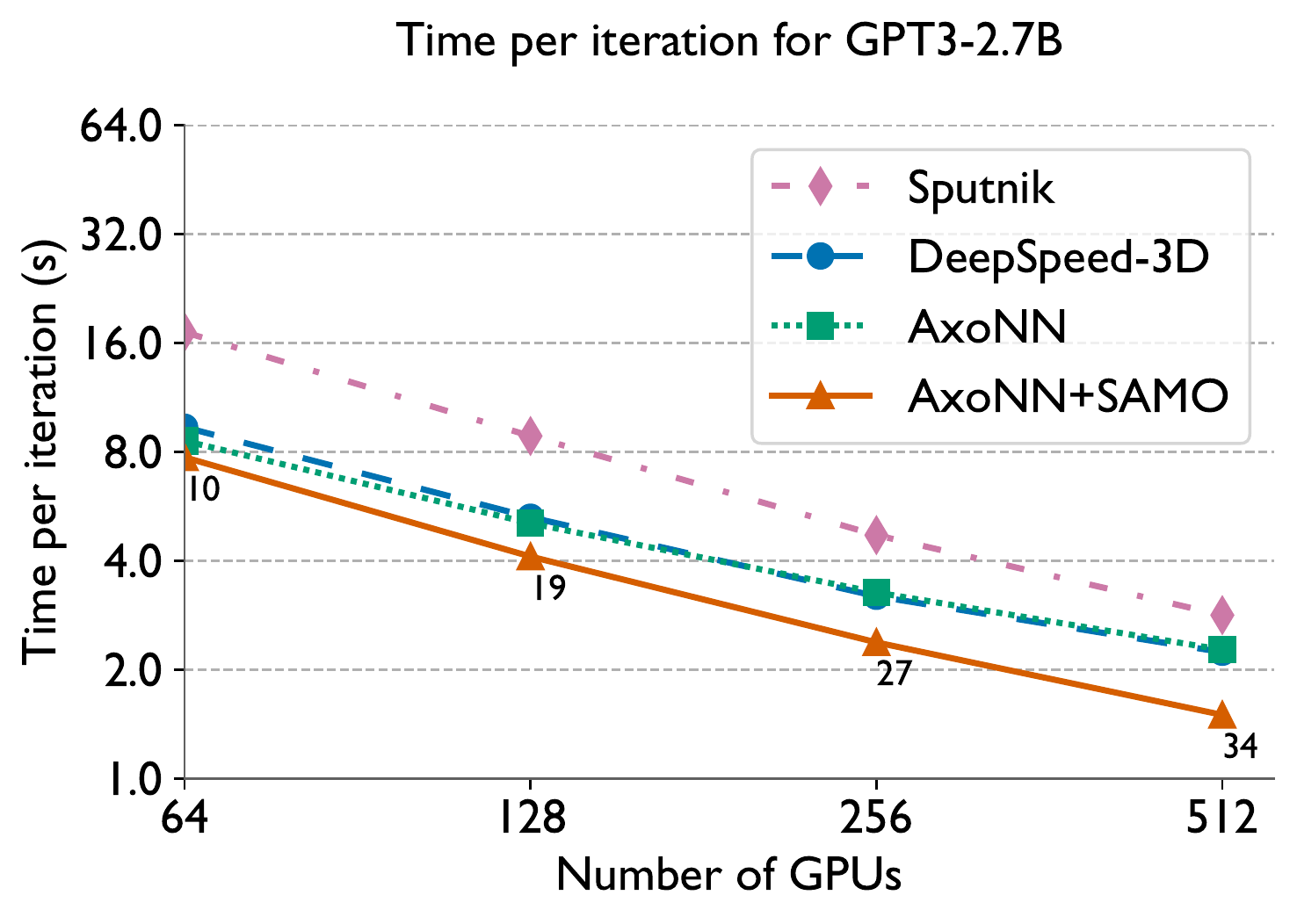}
    \caption{\fix{Time per iteration (batch time) for a strong scaling
  study of GPT-3 XL (left) and GPT-3 2.7B (right) on Summit. We prune the models to a sparsity of 90\% 
  for \axonnplus+\samo and Sputnik (see Table \ref{tab:setup-perf} for batch sizes).
  We annotate \axonnplus+\samo's line with its percentage speedup over \axonn.}}
  \label{fig:hardware-eff-small}
\end{figure*} 

\begin{figure*}[h]
  \centering
    \includegraphics[width=\columnwidth]{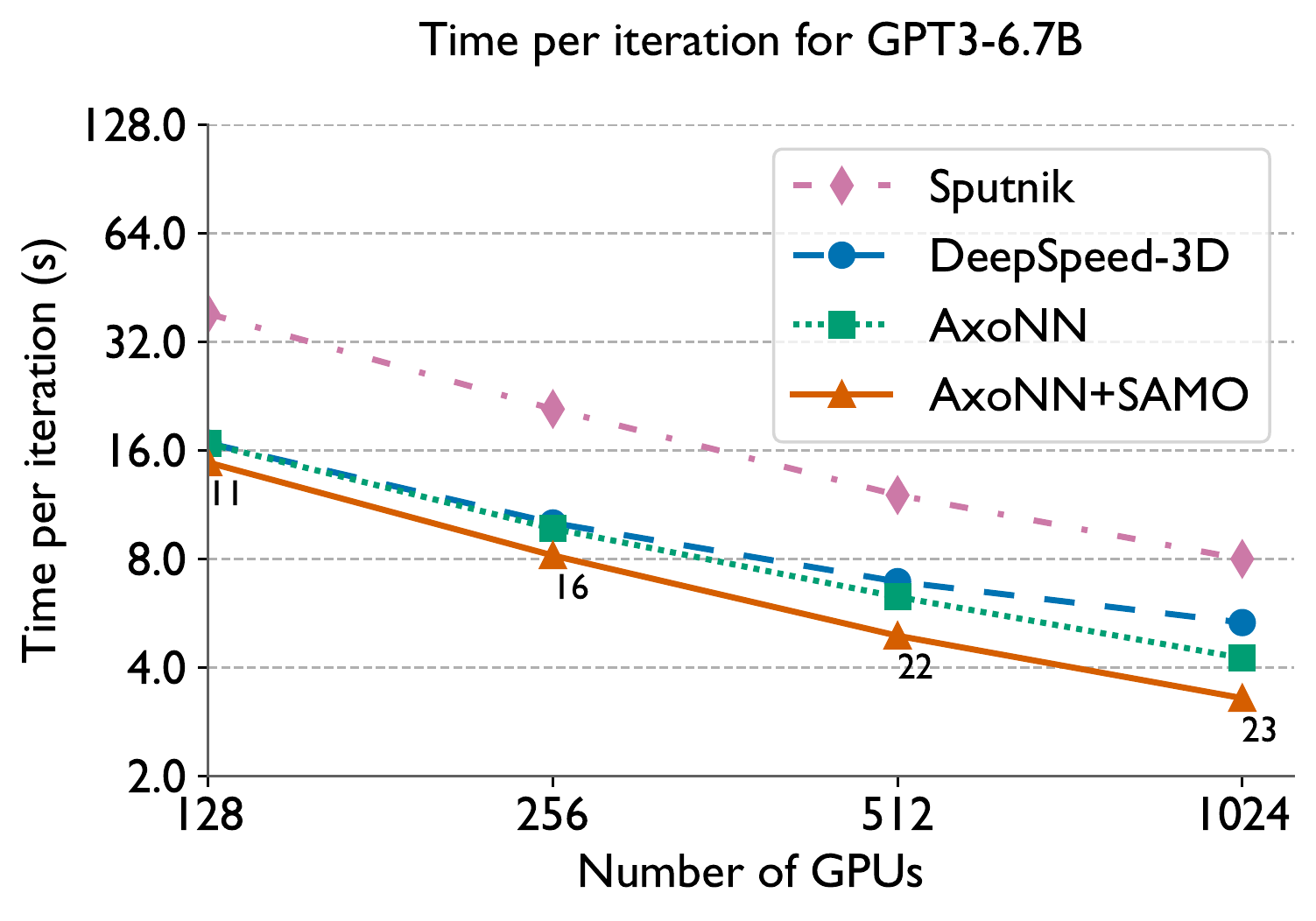}
    \includegraphics[width=\columnwidth]{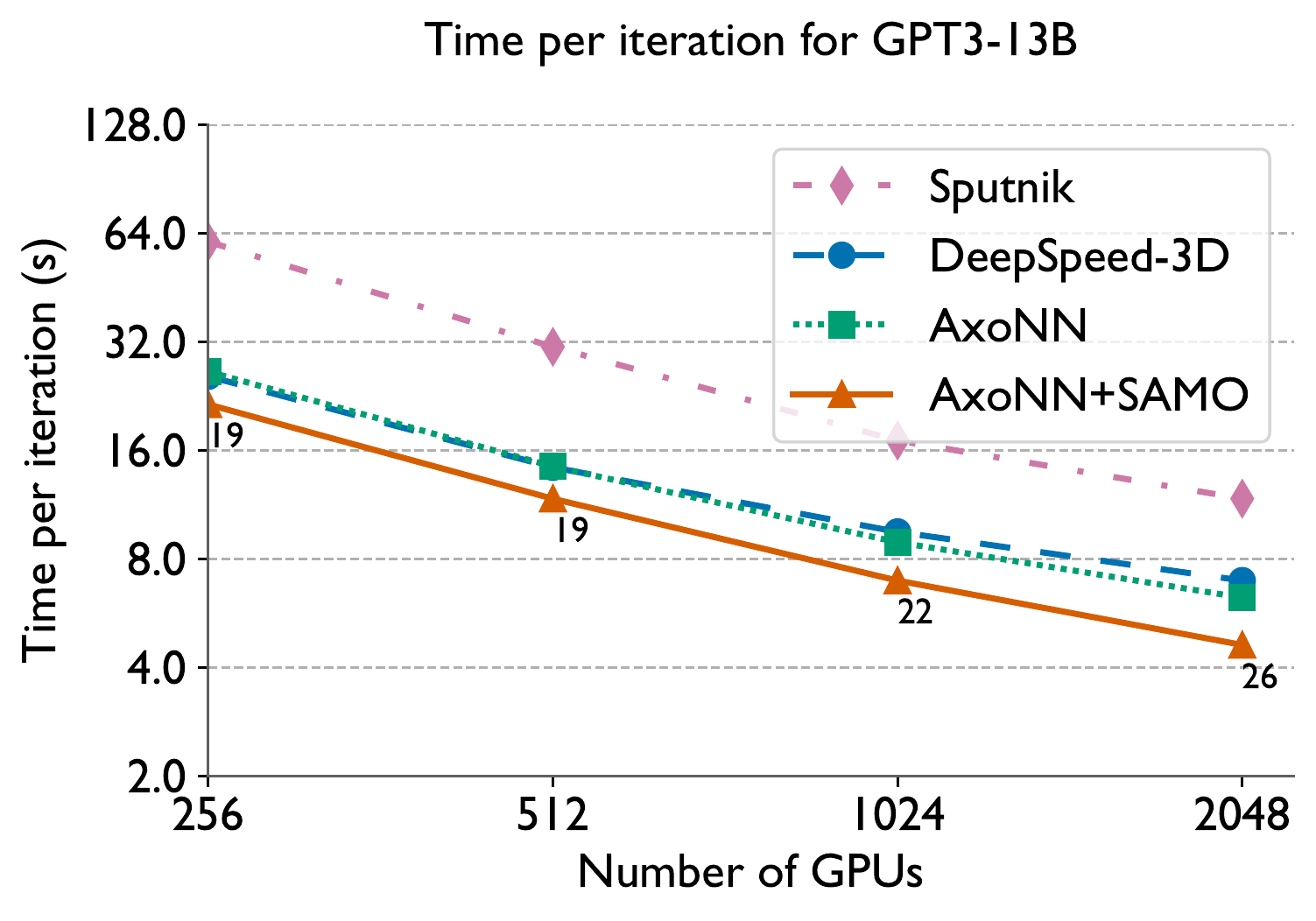}
    \caption{\fix{Time per iteration (batch time) for a strong scaling
  study of GPT-3 6.7B (left) and GPT-3 13B (right) on Summit. We prune the models to a sparsity of 90\% 
  for \axonnplus+\samo and Sputnik (see Table \ref{tab:setup-perf} for batch sizes).
  We annotate \axonnplus+\samo's line with its percentage speedup over \axonn.}}
  \label{fig:hardware-eff-large}
\end{figure*}

Next, we illustrate the results of our strong scaling experiments \fix{on
WideResnet-101 and VGG-19 in Figure \ref{fig:hardware-eff-cnn}}, GPT-3 XL and
GPT-3 2.7B in Figure \ref{fig:hardware-eff-small}, and on GPT-3 6.7B and GPT-3
13B in Figure \ref{fig:hardware-eff-large}. The CNNs used in this study are
nearly 10--100$\times$  smaller than the GPT-3 based models (see Table
\ref{tab:setup-perf}).  Hence, all of \axonn, DeepSpeed-3D and \axonnplus+\samo
are able to run these architectures in a pure data parallel configuration, with
a full copy of the network on each GPU. Thus the only communication here is the
all-reduce on the network gradients. We illustrate these results in Figure
\ref{fig:hardware-eff-cnn}. We observe similar batch times for both \axonn and
DeepSpeed-3D. This is explained by the fact that both these frameworks have
very similar NCCL-based implementations of data parallelism. Our approach
yields speedups of 7--12\% over WideResnet-101 and 18--44\% over VGG-19.  While
both these speedups are significant, \samo seems to benefit the latter
architecture more than the former.  This is because the WideResnet-101
architecture spends nearly 1.5$\times$ more time in the computation phase as
compared to VGG-19. Also, both these models have similar number of parameters
and thus similar communication costs in the data-parallel all-reduce. Thus the
proportion of the batch time spent by the WideResnet-101 architecture in
communication is significantly smaller than VGG-19. Since our approach
optimizes communication, the benefits for WideResnet-101 are smaller than that
of VGG-19. Note that we do not run Sputnik for the CNNs as the library does not
support sparse convolutions.

\fix{Let us now discuss the much larger GPT-3 based neural networks.  These
networks are too large to fit on a single GPU and are thus trained using hybrid
parallelism.} First, we observe that the performance of the sparse matrix
computation library, Sputnik is significantly worse than both of our dense
baselines -- \axonn and DeepSpeed-3D, as well as \axonnplus+\samo
(Figures~\ref{fig:hardware-eff-small} and~\ref{fig:hardware-eff-large}). This
is in spite of the fact that the number of floating point operations computed
by Sputnik is 10\% of the other methods. This is in agreement with our
observations in Figure \ref{fig:sparse-and-dense} for fully connected layers on
a single GPU.  \fix{Thus, \axonnplus+\samo ends up being nearly twice as fast
as Sputnik across all the GPT-3 style neural networks. } It is evident from
Figures \ref{fig:hardware-eff-small} and \ref{fig:hardware-eff-large} that
augmenting \axonn with our optimizations significantly improves its performance
at scale. Our method speeds up the training of GPT-3 XL by 10--47\%, GPT-3 2.7B
by 10--24\%, GPT-3 6.7B by 11--23\% and GPT-3 13B by 19--26\%. The speedups
over DeepSpeed-3D are larger -- 19--51\%, 17--33\%, 12--38\% and 16.4--34\%
respectively for the four models.

\fix{We also present the percentage of peak half precision throughputs obtained
for GPT-3 13B in Table \ref{tab:peak-flops}.} We observe a significant
reduction in the GPU utilization with increasing GPU counts for DeepSpeed-3D
and \axonn. This is a consequence of increasing communication to computation
ratios. For both frameworks, the peak half precision throughput drops to around
20\% at the largest profiled GPU counts. However, with \axonnplus+\samo, we
observe a smaller reduction in hardware utilization, with a peak throughput of
around 30\% for the largest GPU count. This serves as empirical evidence of the
fact that our optimizations indeed decrease the amount of communication in
parallel training.

\begin{table}[h]
  \centering 
  \caption{\fix{Percentage of peak half precision throughput for a strong scaling
  study of GPT-3 13B on Summit (see Table \ref{tab:setup-perf} for batch sizes).
  We prune the models to a sparsity of 90\% for \axonnplus+\samo and Sputnik.
  \label{tab:peak-flops}}}
  \begin{tabular}{lrrrr}
  \toprule
  \# GPUs & Sputnik & DeepSpeed-3D & \axonn & AxoNN+SAMO \\ \midrule
  256            & 18.9    & 44.6         & 43.3   & \textbf{53.4}        \\
  512            & 18.5    & 39.9         & 39.7   & \textbf{48.8}        \\
  1024           & 16.8    & 30.1         & 32.2   & \textbf{41.1}       \\
  2048           & 12.2   & 20.6         & 22.9   &  \textbf{31.0}        \\ 
  \bottomrule
  \end{tabular}
\end{table}

Since our optimizations are geared toward reducing the communication costs of
training, we expect larger improvements over \axonn as the number of GPUs
increase. Again, this is \fix{because} a larger \fix{proportion} of time is
spent in communication as we increase the scale of training. We find our
observations in Figures \ref{fig:hardware-eff-small} and
\ref{fig:hardware-eff-large} to be in agreement with this hypothesis. We indeed
observe the largest speedups for the largest GPU counts, which are 47\% and
34\% for GPT-3 XL and GPT-3 2.7B on 512 GPUs, 23\% for GPT-3 6.7B on 1024 GPUs,
and 26\% for GPT-3 13B on 2048 GPUs.

\subsection{Performance Breakdowns}

\begin{figure}[h]
  \centering
    \includegraphics[width=\columnwidth]{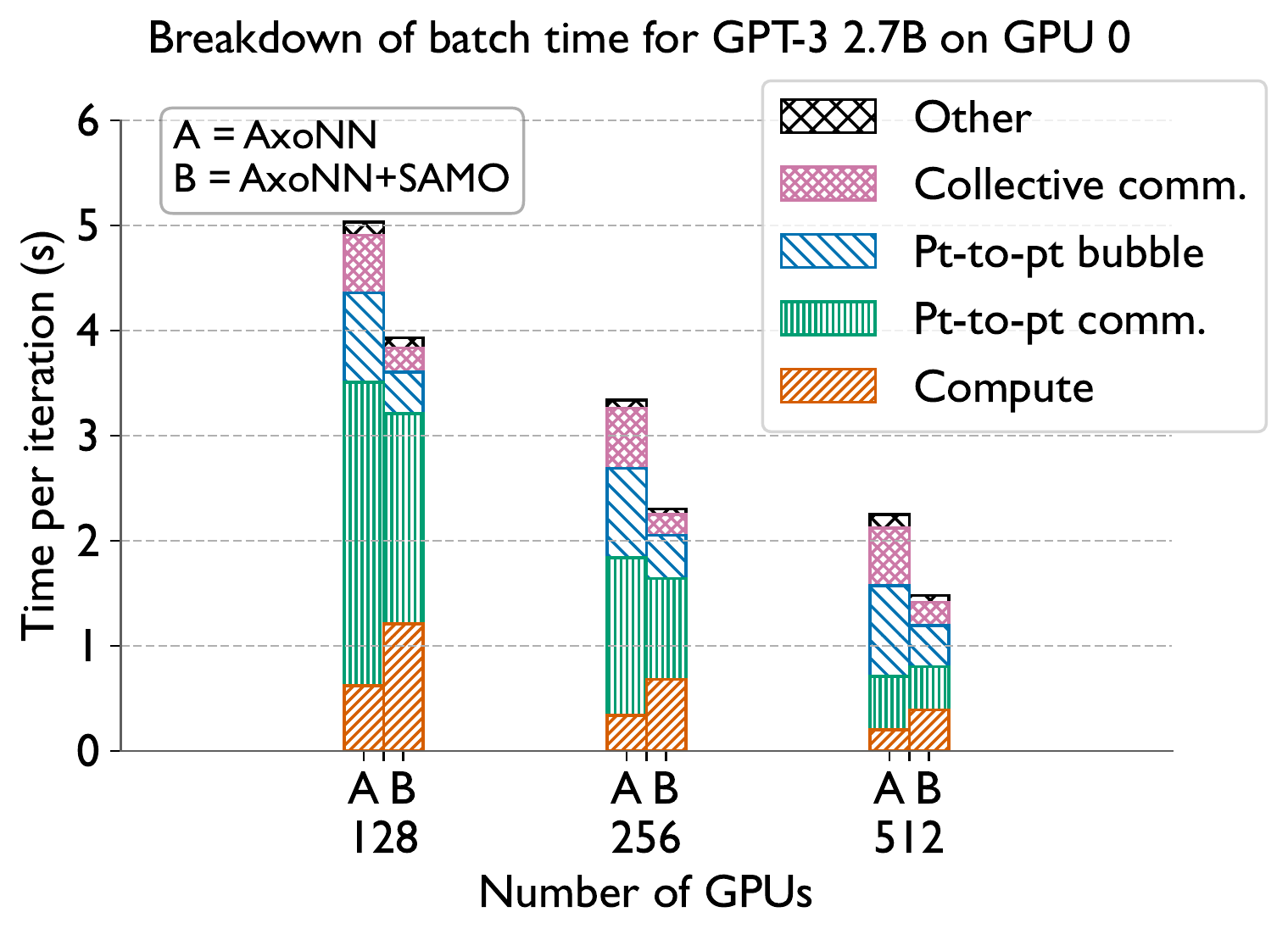}
    \caption{Breakdown of batch time for GPT-3 2.7B on
    Summit. We divide the batch time into its non-overlapping phases --
    computation, point-to-point communication, pipeline bubble (due to
    inter-layer parallelism), and collective communication (due to data
    parallelism). We use the CUDA Event API to profile the cumulative time spent in each of these phases.}
    \label{fig:strong-scale-breakdown}
\end{figure}

To verify that the speedups over \axonn are indeed due to reduction in
communication times, we profile the GPT-3 2.7B model on 128, 256 and 512 GPUs
and provide breakdowns of the batch times in Figure
\ref{fig:strong-scale-breakdown}. We divide the batch time into its
non-overlapping phases, namely the compute (forward and backward pass),
point-to-point communication, pipeline bubble (due to inter-layer parallelism),
and collective communication (due to data parallelism). We use the CUDA Event
API to profile the time spent in each of these phases.

At 128 GPUs, we observe that training is dominated by the point-to-point
communication time. However as the number of GPUs increase, the proportion of
time spent in the point-to-point communication also decreases. Note that this
is in line with Equation \ref{eq:trans}, wherein we showed that the messaging
time is inversely proportional to the number of GPUs.

We observe that the primary reason for \axonnplus+\samo's improvement over
\axonn on 128 GPUs is due to a speedup in the point-to-point communication
times. The absolute reduction in this time is 18\% of \axonn's batch time. The
improvements in the collective and pipeline bubble times account for 6\% and
9\% of \axonn's batch time. Thus for smaller GPU counts, we conclude that
\axonnplus+\samo provides speedups primarily because of the improvements in the
point-to-point communication times. The difference in the compute times is the
overhead of compressing the parameter gradients at every backward pass (See
Section \ref{sec:train-samo}). The overhead accounts for 12\% of \axonn's batch
time and is significantly overcome by the 33\% (18+6+9) improvement in the
total communication time. We think that these overheads can be reduced by
kernel level optimizations such as fusing the compression operation with the
backward pass kernels. However, this is out of the scope of our current work.

At 256 GPUs, the point-to-point communication time is still dominant but not as
much as 128 GPUs.  In this case, the improvement in the point-to-point
communication time accounts for a 16.17\% of \axonn's batch time. As compared
to 128 GPUs, the improvements in the bubble and collective communication times
account for a significantly larger proportion of \axonn's batch time - 13.17\%
and 11.08\%. The overhead in this case is 10.18\% of the total batch time.

At 512 GPUs, we notice a very minor reduction in the point-to-point
communication time. The reduction in the bubble and collective communication
time account for 15\% and 21\% of \axonn's batch time respectively. The
reduction in the point-to-point communication only improves the batch times by
4\%. In this case, the overhead of compressing gradients is 8\% of \axonn's
batch time, which is again overcome by 40\% (15+21+4) improvement in the total
communication times.

\section{Conclusion}
It is well known that recent magnitude-based pruning approaches can lead to
significant pruning of neural networks without reducing statistical
efficiency~\cite{frankle2018the, You2020Drawing}.  However, to the best of our
knowledge, no prior work has attempted to exploit neural network pruning for
improving the hardware efficiency of parallel neural network training. The
primary deterrent to doing this is the sparse nature of the pruned subnetworks,
which results in inefficient hardware performance. 

In this work, we presented Sparsity-aware Memory Optimization (\samo), a novel
method that exploits the aforementioned parameter pruning algorithms to
significantly reduce the memory consumption of neural network training while
not sacrificing performance. We then demonstrated how these memory savings can
be utilized to significantly improve the communication performance of two
popular algorithms for parallel deep learning, data and inter-layer
parallelism. We conducted strong scaling experiments \fix{on two convolution
neural networks,} and large GPT-style language models with 1.3 billion to 13
billion parameters proposed by Brown et al.~\cite{gpt-3} in their seminal work
on the GPT-3 architecture.  In our experiments, we consistently observed
significant improvements over two highly scalable parallel deep learning
frameworks -- \axonn and DeepSpeed-3D and a state-of-the-art sparse matrix
computation library called Sputnik.

\section*{Acknowledgment}
This work was supported by funding provided by the University of Maryland
College Park Foundation. The authors thank Shu-Huai Lin for his help in
conducting initial experiments that led to this research. This research used
resources of the Oak Ridge Leadership Computing Facility at the Oak Ridge
National Laboratory, which is supported by the Office of Science of the
U.S.~Department of Energy under Contract No.~DE-AC05-00OR22725.

\IEEEtriggeratref{47}
\bibliographystyle{IEEEtran}
\bibliography{./bib/cite,./bib/pssg}

\end{document}